\documentclass[runningheads]{llncs}

\usepackage[title]{appendix}

\usepackage[dvipsnames]{xcolor}
\usepackage[colorlinks=true,urlcolor=blue]{hyperref}
\usepackage[pdftex]{graphicx}
\usepackage{booktabs}
\usepackage[caption=false,font=footnotesize]{subfig}
\usepackage{tabularx}
\usepackage{wrapfig}
\usepackage{lipsum}
\usepackage{amsmath}
\graphicspath{{./images/}}

\newcommand{\fac}[1]{\textcolor{Sepia}{#1}}

\begin{document}

\title{An Open-Source Multi-Goal Reinforcement Learning Environment for Robotic Manipulation with Pybullet\thanks{The authors thank the China Scholarship Council (CSC) for financially supporting Xintong Yang in his PhD programme (No. 201908440400).}}

\titlerunning{This paper was submitted to Taros 2021}

\author{Xintong Yang\inst{1}\orcidID{0000-0002-7612-614X} \and
Ze Ji\inst{1}\orcidID{0000-0002-8968-9902}\thanks{Corresponding author} \and
Jing Wu\inst{2}\orcidID{0000-0001-5123-9861} \and
Yu-Kun Lai\inst{2}\orcidID{0000-0002-2094-5680}}

\authorrunning{X. Yang et al.}

\institute{Centre for Artificial Intelligence, Robotics and Human-Machine Systems (IROHMS), School of Engineering, Cardiff University, Cardiff, UK \\\email{\{yangx66, jiz1\}@cardiff.ac.uk} \and
School of Computer Science and Informatics, Cardiff University, Cardiff, UK \\\email{\{wuj11, laiy4\}@cardiff.ac.uk}}

\maketitle

\begin{abstract}
This work re-implements the OpenAI Gym multi-goal robotic manipulation environment, originally based on the commercial Mujoco engine, onto the open-source Pybullet engine. By comparing the performances of the Hindsight Experience Replay-aided Deep Deterministic Policy Gradient agent on both environments, we demonstrate our successful re-implementation of the original environment. Besides, we provide users with new APIs to access a joint control mode, image observations and goals with customisable camera and a built-in on-hand camera. We further design a set of multi-step, multi-goal, long-horizon and sparse reward robotic manipulation tasks, aiming to inspire new goal-conditioned reinforcement learning algorithms for such challenges. We use a simple, human-prior-based curriculum learning method to benchmark the multi-step manipulation tasks. Discussions about future research opportunities regarding this kind of tasks are also provided.

\end{abstract}
\section{Introduction}
Due to the difficulties of reinforcement learning in real-world environments \cite{dulac2019challenges}, developing simulation environments for robotic manipulation tasks becomes increasingly important. In addition to the requirement of being realistic, such simulation is also required to be efficient in generating synthetic data for training deep reinforcement learning (DRL) agents. Currently, the most popular physics engines in DRL research are Mujoco \cite{plappert2018multi,tassa2018deepmind,todorov2012mujoco} and Pybullet \cite{coumans2019bullet,delhaisse2020pyrobolearn,shenigibson}. Mujoco is known to be more efficient than Pybullet \cite{erez2015simulation}, but it is not open-sourced. 

The cost of a Mujoco institutional license is at least \${3000} per year \cite{muojoco2018page}, which is often unaffordable for many small research teams, especially when a long-term project depends on it. To promote wider accessibility to such resource and support DRL research in robot arm manipulations, we introduce an open-source simulation software, \textbf{PMG}, \textbf{P}ybullet-based, \textbf{M}ulti-goal, \textbf{G}ym-style \cite{brockman2016openai}. It is written in Python, the most popular language in recent machine learning research\footnote{The source codes are available at \url{https://github.com/IanYangChina/pybullet_multigoal_gym}.}.

The manipulation tasks proposed by \cite{Andrychowicz2017,plappert2018multi} focus on goal-condition reinforcement learning (GRL) in sparse reward scenarios. GRL aims to train a policy that behaves differently when given different goals, for example, picking up different objects. While in sparse reward cases, the agent only receives a reward signal when a goal is achieved. This is motivated by the fact that providing task completion information is often easier and less biased than hand-designing a behaviour-specific reward function for most real-world robotic tasks \cite{dulac2019challenges}.

\begin{figure}[t]
    \centering	
	\subfloat[\label{fig:single-step-tasks}From left to right: KukaReach, KukaPickAndPlace, KukaPush and KukaSlide.]{
	\includegraphics[width=\textwidth]{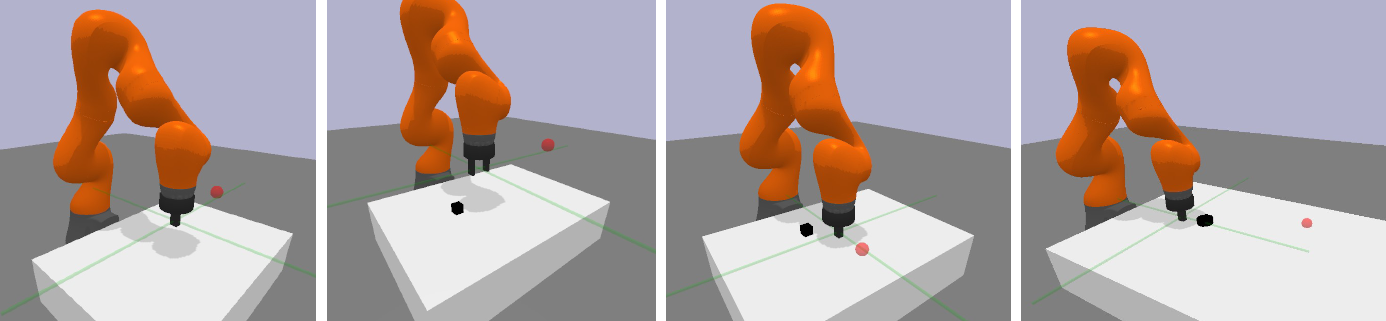}}
    \\[-0.1ex]
    \subfloat[\label{fig:multi-step-tasks}From left to right: BlockRearrange, ChestPush, ChestPickAndPlace, BlockStack]{
    \includegraphics[width=\textwidth]{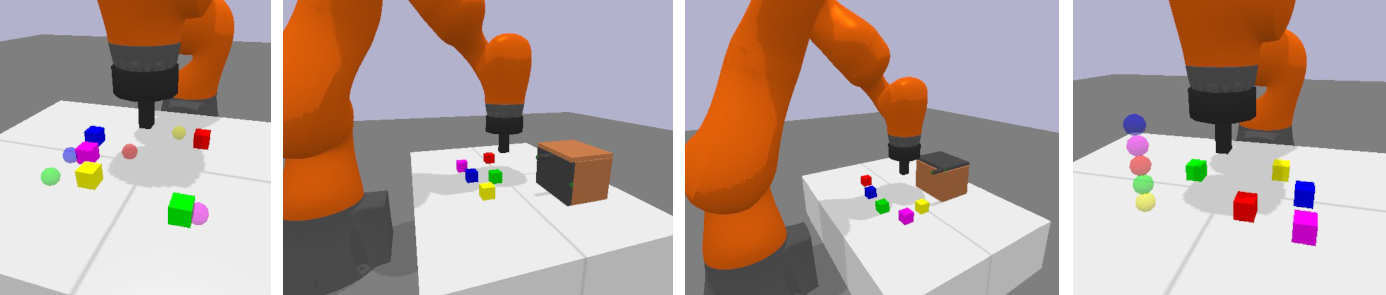}}
	\caption{The robotic arm manipulation tasks. (a) Single-step tasks, reproduced from the original OpenAI Gym multi-goal manipulation tasks \cite{Andrychowicz2017,Lillicrap2015} (described in section~\ref{single-step-task-description}). (b) Multi-step tasks (described in section~\ref{multi-step-task-description}).\label{fig:tasks}}
	\vspace{-15pt}
\end{figure}

We implement the four basic tasks (Fig.~\ref{fig:single-step-tasks}) proposed in \cite{Andrychowicz2017} using Pybullet and reproduce the performances achieved by the Deep Deterministic Policy Gradient (DDPG) algorithm with Hindsight Experience Replay (HER) \cite{Andrychowicz2017,Lillicrap2015}. 

In addition, we further propose a set of new tasks that focus on multi-step manipulations in longer horizon with sparse rewards (Fig.~\ref{fig:multi-step-tasks}). To improve readability, the original set of tasks is named `\textsf{single-step tasks}' and the new set of tasks is named `\textsf{multi-step tasks}'. The \textsf{multi-step tasks} are developed with the aim to inspire new learning algorithms that can handle tasks where the reward signals only appear near the end of the task horizon \cite{dulac2019challenges,yang2021hierarchical}. Beside the delayed rewards, these tasks also require multiple steps to complete, and some of the steps are strongly dependent. For example, a block cannot be placed into a chest unless the chest is opened. This characteristic requires a learning algorithm to reason about the relationships between steps.

To facilitate comparison in future research, we benchmarked the performances on the four \textsf{multi-step tasks} by training the aforementioned DDPG-HER agent \cite{Andrychowicz2017} with a simple human prior-based curriculum. Potential research directions in this regard are also discussed. To sum up, our contributions in this article are:

\begin{itemize}
\item[$\bullet$] Reproducing the multi-goal robotic arm manipulation tasks \cite{plappert2018multi} using Pybullet, making it freely accessible.
\item[$\bullet$] Reproducing the Hindsight Experience Replay performances \cite{Andrychowicz2017} on the Pybullet-based environments.
\item[$\bullet$] Proposing a set of new environments for multi-goal multi-step long-horizon sparse reward robotic arm manipulations.
\item[$\bullet$] Benchmarking the \textsf{multi-step tasks} and proposing future research opportunities.
\end{itemize}

The rest of this paper includes the details of the proposed environments and programming APIs (section~\ref{sec:envs}); the reproduction results of the DDPG-HER agent on the \textsf{single-step tasks}, the benchmark results of the \textsf{multi-step tasks} and discussions of challenges and future research (section~\ref{sec:results}); and finally the conclusion (section~\ref{sec:conclusion}).

\section{Environment}\label{sec:envs}
\subsection{Single-step tasks}
As shown in Fig.~\ref{fig:single-step-tasks}, the \textsf{single-step tasks} are:
\begin{itemize}\label{single-step-task-description}
\item[$\bullet$] \textsf{KukaReach}, where the robot needs to move the gripper tip to a goal location.
\item[$\bullet$] \textsf{KukaPickAndPlace}, where the robot needs to pick up the block and move it to a goal location\footnote{In training, the PickAndPlace goals are generated either on the table surface or in the air, with even probability, as suggested by \cite{Andrychowicz2017}}.
\item[$\bullet$] \textsf{KukaPush}, where the robot needs to push the block to a goal location on the table surface.
\item[$\bullet$] \textsf{KukaSlide}, where the robot needs to push the cylinder bulk with a force such that the bulk slides to a goal location that is unreachable by the robot.
\end{itemize}

Different from the original environments, which use a Fetch robot, we use a Kuka IIWA 14 LBR robot arm equipped with a simple parallel jaw gripper. This does not affect training as only Cartesian space control (gripper movement and finger width) are used in the original tasks. We plan to support more robot arms in the future. 

In addition to the gripper frame control mode, our environments also support joint space control, which results in a $7$ dimensional action space for the \textsf{KukaReach} and \textsf{KukaPush} tasks and an $8$ dimensional one for the other two tasks (with one extra dimension for controlling the gripper finger width). Such a control mode has been largely ignored in most DRL-based manipulation works, possibly due to its high dimensionality. However, this control mode is important in scenarios that involve collision avoidance. A manipulation policy should not only consider end-effector control, but also learn to control each joint more explicitly when the surroundings are crowded by objects or other agents, e.g., humans. We leave the design of tasks for this specific direction to future work.

The tasks provide two reward functions. The dense reward function uses the negative Euclidean distance between the achieved and desired goals. The sparse reward function gives a reward of 0 when a goal is achieved and -1 everywhere else. We further provide RGB-D images as an optional observation representation. Users can easily define different camera view-points for rendering observations and goals.

Note that, we did not change the design of these four tasks, but reproduce them using a different physics engine.  For more details of the task, such as the state and the action spaces, we refer the readers to the original paper \cite{plappert2018multi}. The APIs and programming style are slightly different and are described in section~\ref{sec:programming}.

\subsection{Multi-step tasks}\label{sec:multi-step-tasks}

Fig.~\ref{fig:multi-step-tasks} visualises the four challenging \textsf{multi-step tasks} developed by the authors, aiming at sparse reward long-horizon manipulations. Briefly, they are:
\begin{itemize}\label{multi-step-task-description}
\item[$\bullet$] \textsf{BlockRearrange}, where the robot needs to push the blocks to random positions. Gripper fingers are blocked in this task.
\item[$\bullet$] \textsf{ChestPush}, where the robot needs to first open the sliding door (in black colour) of the chest and then push the blocks into the chest. Gripper fingers are blocked in this task. 
\item[$\bullet$] \textsf{ChestPickAndPlace}, where the robot needs to first open the sliding door (in black colour) of the chest and then pick and drop the blocks into the chest.
\item[$\bullet$] \textsf{BlockStack}, where the robot needs to stack the blocks into a tower in a given order that is randomly chosen.
\end{itemize}

These tasks require the robot to learn different combinations of behaviours and provide different numbers of step dependencies. For example, the \textsf{BlockStack} task has more dependent steps with the increase of the number of blocks to be stacked. The complexity of these tasks increases with more dependent steps and blocks, as shown in Table.~\ref{tab:multi-step-task-summary}. Moreover, the number of blocks involved in a task affects its task horizon, and thus its exploration difficulty. Detailed task information is provided in \href{https://github.com/IanYangChina/taros2021codes/blob/master/taros20210308_supplementary.pdf}{supplementary material} section 1.

With the challenge of sparse reward in mind, the extreme case of these tasks is that the environment only gives a task completion signal (e.g., a reward of 0) when the ultimate goal (e.g., all the blocks are stacked) is achieved, and provides a reward of -1 everywhere else. In this case, the task is extremely difficult for any naive reinforcement learning algorithm, even the one with hindsight experience replay (see section~\ref{sec:multi-step-result}). This is because the reinforcement learning agent has an extremely low probability of seeing a meaningful reward value. Compared to the \textsf{single-step tasks}, which only feature the sparse reward problem in a short task horizon, these \textsf{multi-step tasks} can be used to investigate more difficult problems, such as
\begin{itemize}
\item[$\bullet$] How to explore efficiently for multi-step tasks with sparse and delayed rewards?
\item[$\bullet$] How to represent and learn the dependencies among task steps?
\item[$\bullet$] How can ideas such as curriculum learning, option discovery and hierarchical learning help in these tasks?
\end{itemize}

\begin{table}[t]
\caption{\textsf{Multi-step tasks summary}\label{tab:multi-step-task-summary}}
\begin{center}
\begin{tabularx}{\textwidth}{llll}
\toprule
\textbf{Task} & \textbf{Needed behaviours} \ & \textbf{Step dependency} \ & \textbf{Num. of blocks}\\
\hline
\textsf{BlockReaarange} \ & pushing & $0$ & $2$ to $5$\\
\textsf{ChestPush} & pushing & $1$ & $1$ to $5$\\
\textsf{ChestPickAndPlace} & pushing, picking, dropping & $1$ & $1$ to $5$\\
\textsf{BlockStack} & pushing, picking, placing & $\geq 2$ & $2$ to $5$\\
\bottomrule
\end{tabularx}
\end{center}
\vspace{-15pt}
\end{table}

One possible research direction for these problems is to create a curriculum that provides the learning algorithm with goals starting from easy to difficult \cite{manela2020curriculum,narvekar2020curriculum}. In this paper, we design a human-prior based curriculum for the \textsf{multi-step tasks}. It simply generates goals that require increasing time horizons to achieve, e.g., from stacking two blocks to five. However, the results show that such a simple curriculum is not efficient enough for longer horizon tasks (see section~\ref{sec:multi-step-result}). To tackle these problems, more efficient methods need to be developed. Section~\ref{sec:challenges} provides more discussion on future research opportunities.

\subsection{APIs and Programming style}\label{sec:programming}
In OpenAI Gym, users create environment instances by specifying a unique task ID pre-registered in the package \cite{brockman2016openai,plappert2018multi}. In contrary, we provide users with an API to make environments more intuitively. As shown in \hyperref[code:env-instance]{Code 1}, the \texttt{make\_env(...)} function provides arguments to setup a specific environment instance. \href{https://github.com/IanYangChina/taros2021codes/blob/master/taros20210308_supplementary.pdf}{Supplementary material} section 2 provides a detailed explanation of these arguments. Currently, only eight tasks are prepared, including four \textsf{single-step tasks} and four \textsf{multi-step tasks}. 

\begin{figure}[h]
	\vspace{-15pt}
    \centering
	\subfloat[\label{fig:obs-scene}Scene]{
	\includegraphics[width=0.24\textwidth]{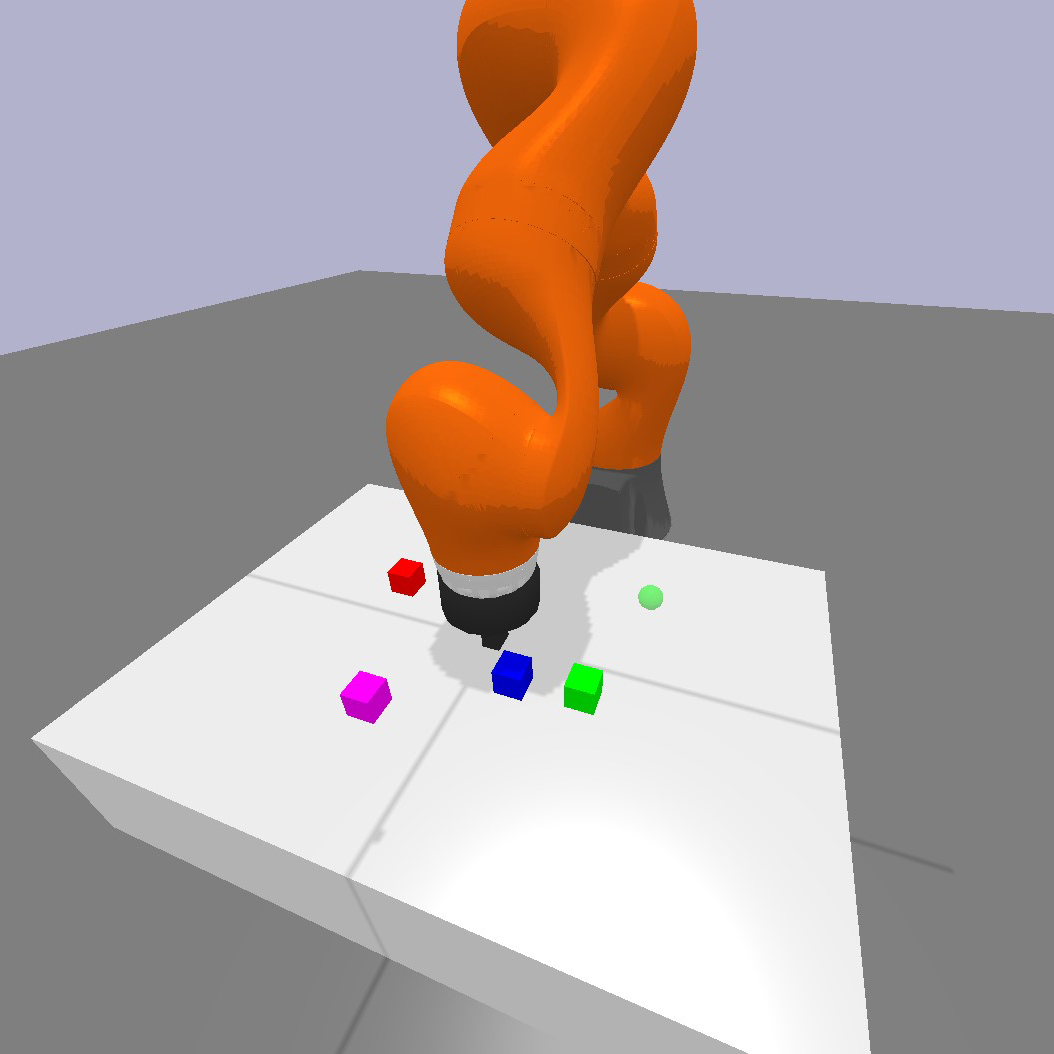}}
    \hfil
    \subfloat[\label{fig:cam-1}Camera 1]{
    \includegraphics[width=0.24\textwidth]{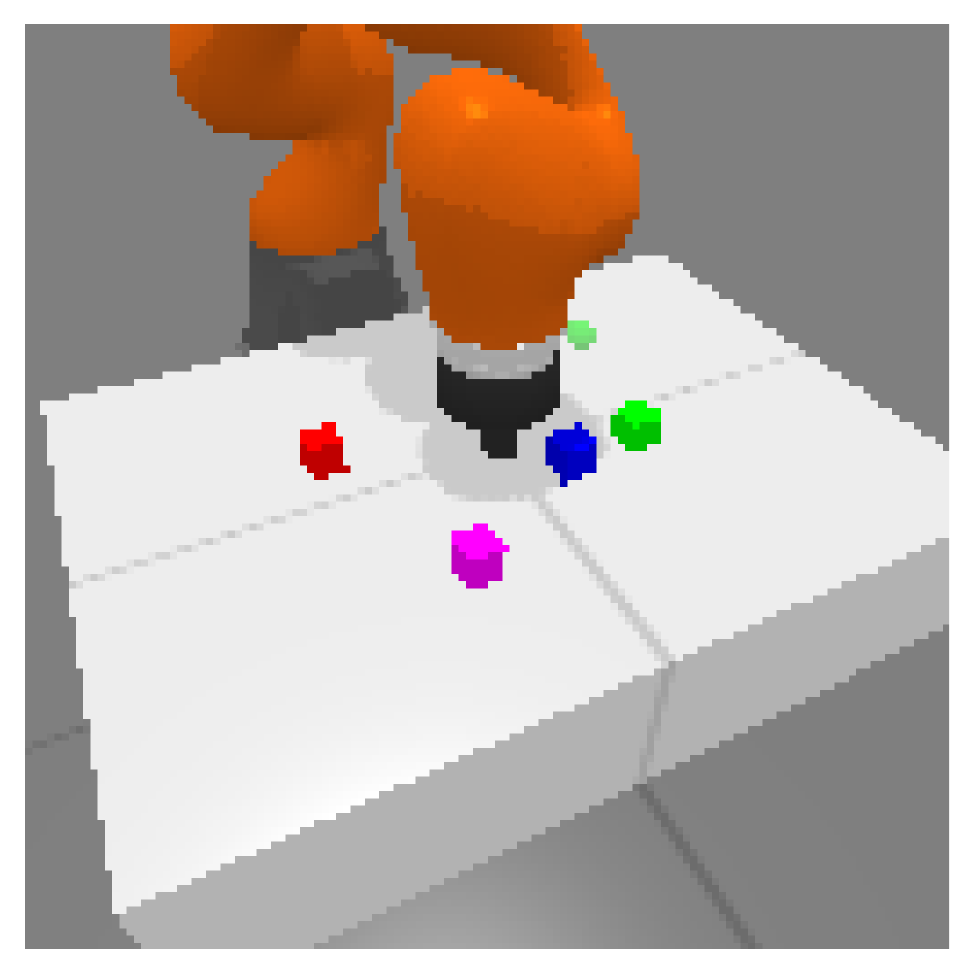}}
	\hfil
	\subfloat[\label{fig:cam-2}Camera 2]{
	\includegraphics[width=0.24\textwidth]{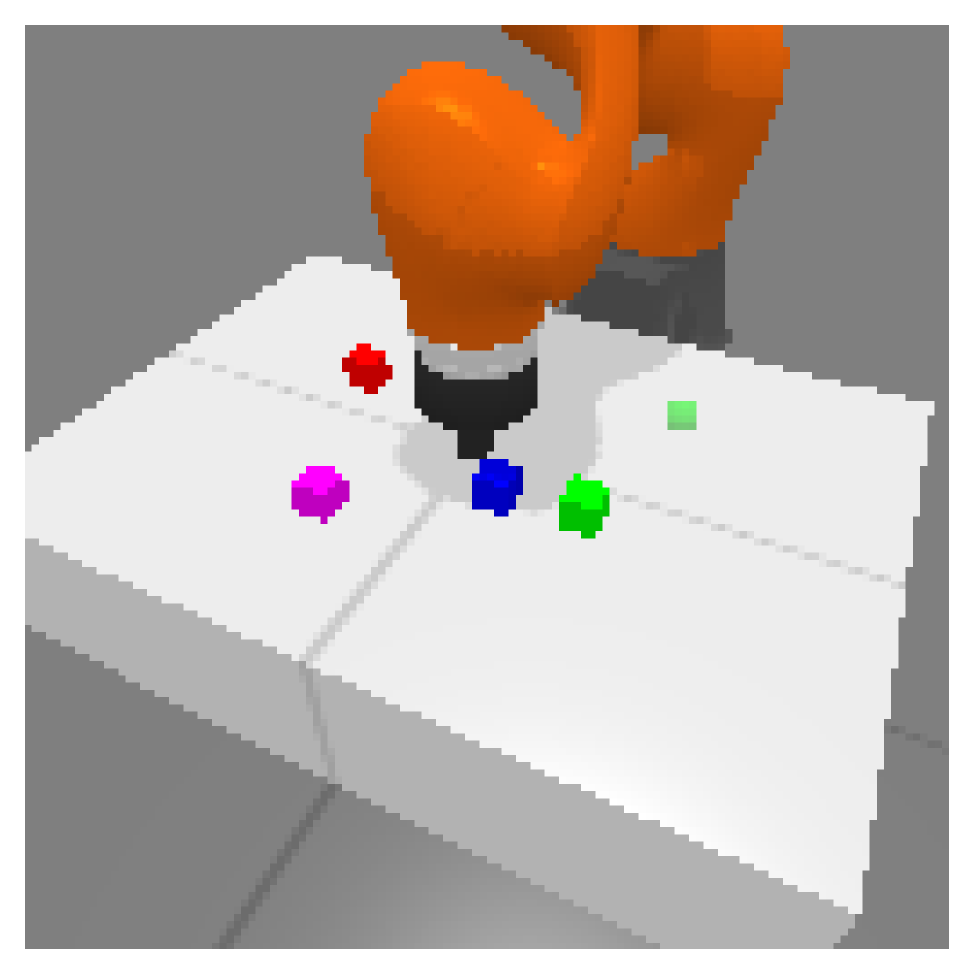}}
    \hfil
    \subfloat[\label{fig:cam-on-hand}On-hand camera]{
    \includegraphics[width=0.24\textwidth]{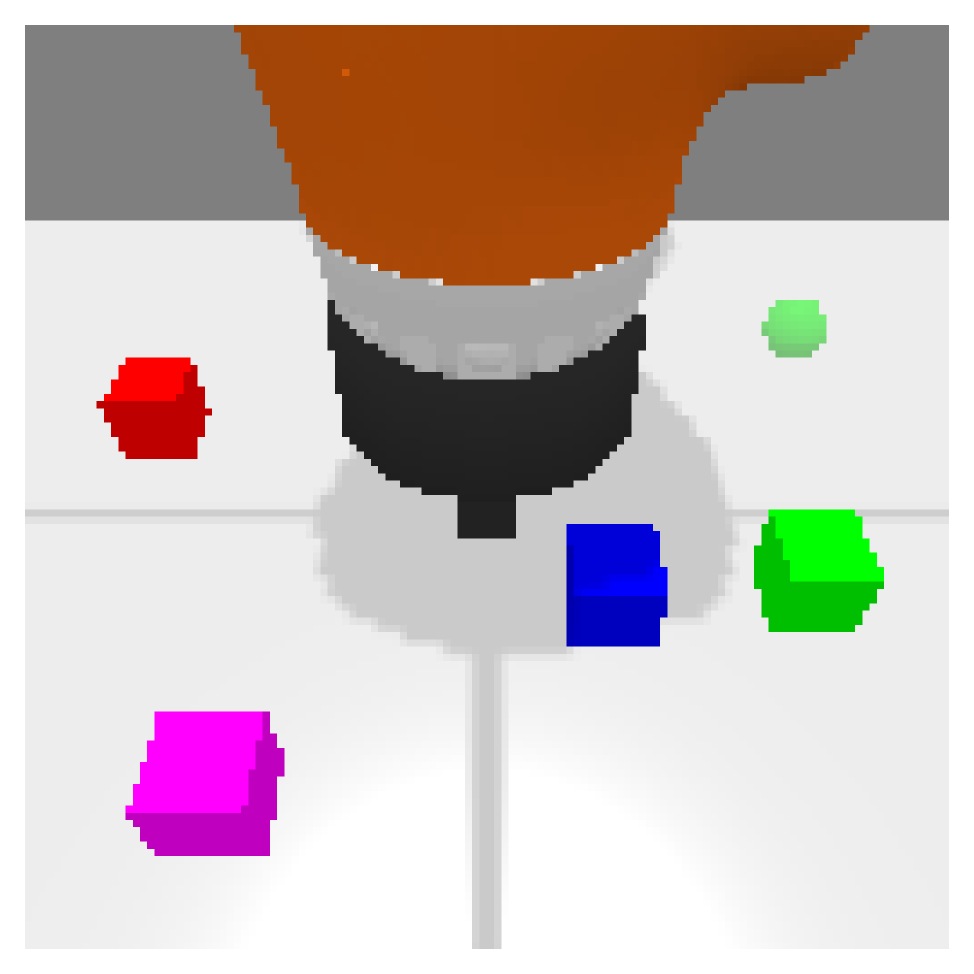}}
	\caption{Images rendered using the two cameras defined in \href{https://github.com/IanYangChina/taros2021codes/blob/master/taros20210308_supplementary.pdf}{supplementary material} Code 2 and the built-in on-hand camera.\label{fig:camera-rendering}}
	\vspace{-15pt}
\end{figure}

We provide an argument to activate image observations and goals, while the original Gym environment requires users to rewrite some of the code to achieve this. In addition, users can easily customise cameras for observation or goal images by defining a list of Python dictionaries and passing it to the \texttt{camera\_setup} argument. An example is given in \href{https://github.com/IanYangChina/taros2021codes/blob/master/taros20210308_supplementary.pdf}{supplementary material} Code 2. Intuitively, the setup example defines two cameras, and in  \hyperref[code:env-instance]{Code 1} they are used for capturing observation and goal images respectively, by setting the \texttt{cam\_id} arguments to $0$ and $1$. Alternatively, users can pass $-1$ to the \texttt{cam\_id} arguments, activating an on-hand camera looking at the gripper tip position. Fig.~\ref{fig:camera-rendering} shows a scene and three images rendered with the above-mentioned cameras.

Except for the codes that create an environment instance, other user APIs are kept the same as the original multi-goal Gym environment package. In our experiments, the code of training the DDPG-HER agent needs no change from Mujoco to Pybullet, and we successfully reproduce the performances as shown in section~\ref{sec:single-step-result}.

\begin{table}[t]
\begin{center}
\begin{tabular}{l}
\toprule
\textbf{Code 1}\label{code:env-instance} Create an environment instance\\
\midrule
\# Original OpenAI Gym style\\
import gym\\
env = gym.make(``FetchReach-v0'')\\
\# Our style\\
import pybullet\_multigoal\_gym as pmg\\
env = pmg.make\_env(\\
\ \ \ \ \ \ \# task args\\
\ \ \ \ \ \ \texttt{\fac{task}}=`block\_rearrange', \texttt{\fac{joint\_control}}=False, \texttt{\fac{num\_block}}=2, \texttt{\fac{render}}=False,\\
\ \ \ \ \ \ \texttt{\fac{binary\_reward}}=True, \texttt{\fac{max\_episode\_steps}}=50, \texttt{\fac{distance\_threshold}}=0.05\\
\ \ \ \ \ \ \# image observation args\\
\ \ \ \ \ \ \texttt{\fac{image\_observation}}=False, \texttt{\fac{depth\_image}}=False, \texttt{\fac{goal\_image}}=False,\\
\ \ \ \ \ \ \texttt{\fac{visualize\_target}}=True,\\
\ \ \ \ \ \ \texttt{\fac{camera\_setup}}=camera\_setup, \texttt{\fac{observation\_cam\_id}}=0, \texttt{\fac{goal\_cam\_id}}=1,\\
\ \ \ \ \ \ \# curriculum args\\
\ \ \ \ \ \ \texttt{\fac{use\_curriculum}}=True, \texttt{\fac{num\_goals\_to\_generate}}=1e6)\\
\\
\# Interaction loop\\
obs = env.reset()\\
while True:\\
\ \ \ \ \ \ action = env.action\_space.sample()\\
\ \ \ \ \ \ obs, reward, done, info = env.step(action)\\
\ \ \ \ \ \ if done:\\
\ \ \ \ \ \ \ \ \ \ \ \ obs = env.reset()\\
\bottomrule
\end{tabular}
\end{center}
\vspace{-15pt}
\end{table}

\section{Benchmark and Discussion}\label{sec:results}
In section~\ref{sec:single-step-result}, we reproduced the Hindsight Experience Replay (HER) \cite{Andrychowicz2017} on the \textsf{single-step tasks} to demonstrate the success of the transfer from the Mujoco-based environments to ours. More specifically, we trained a DDPG agent using the `future' goal-relabelling strategies, with the same hyperparameters and design proposed in \cite{Andrychowicz2017}, except that we did not use distributed training. In addition, we also trained the same agent on the \textsf{single-step tasks} with joint control. 

Section~\ref{sec:multi-step-result} shows the results of training the DDPG-HER agent on the \textsf{multi-step tasks}. The results serve as a benchmark for future studies. Section~\ref{sec:challenges} provides challenges and future research opportunities.

The Pytorch implementation of the algorithm is available 
\href{https://github.com/IanYangChina/DRL_Implementation/blob/master/drl_implementation/agent/continuous_action/ddpg_goal_conditioned.py}{here}. The experiment scripts are available \href{https://github.com/IanYangChina/taros2021codes}{here}. All experiments of this article were run on Ubuntu 16.04 on a workstation with an Intel i7-8700 CPU and an Nvidia RTX-2080Ti GPU. All performance statistics are averaged from $4$ runs with different random seeds.

\subsection{Reproducing Hindsight Experience Replay on Single-step tasks}\label{sec:single-step-result}
\begin{wrapfigure}{r}{4cm}
\vspace{-25pt}
\includegraphics[width=4cm]{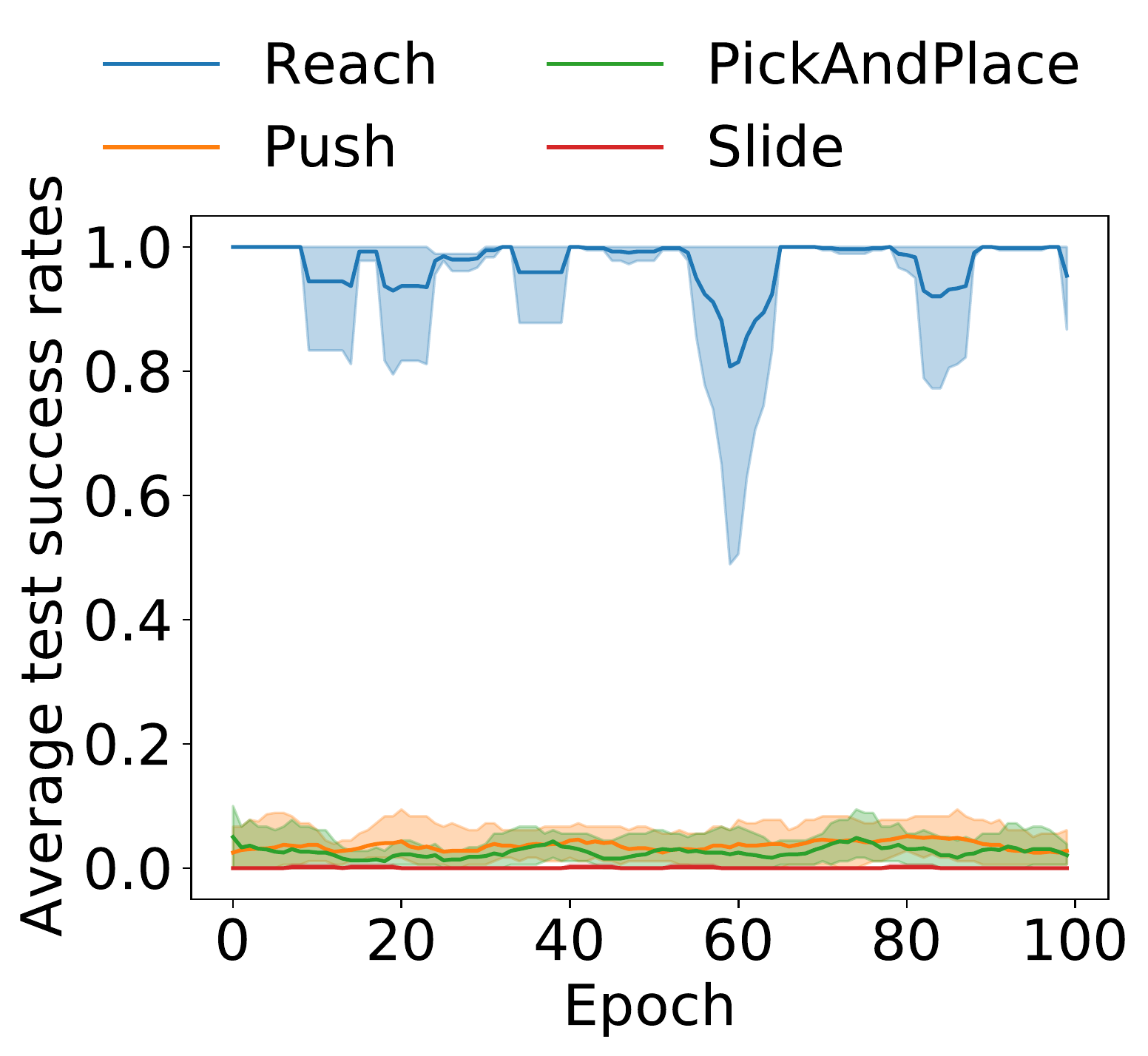}
\caption{DDPG-HER performances on Joint space control tasks.}\label{fig:joint-ctrl}
\vspace{-20pt}
\end{wrapfigure} 

\begin{figure}[h]
	\vspace{-5pt}
    \centering
    \subfloat{
	\includegraphics[width=0.9\textwidth]{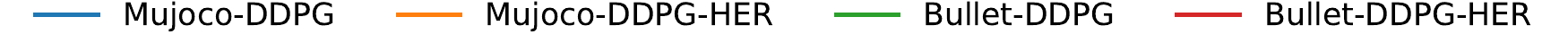}}\\[-2ex]
    \addtocounter{subfigure}{-1}
	\subfloat[\label{fig:reach}Reach]{
	\includegraphics[height=0.208\textwidth]{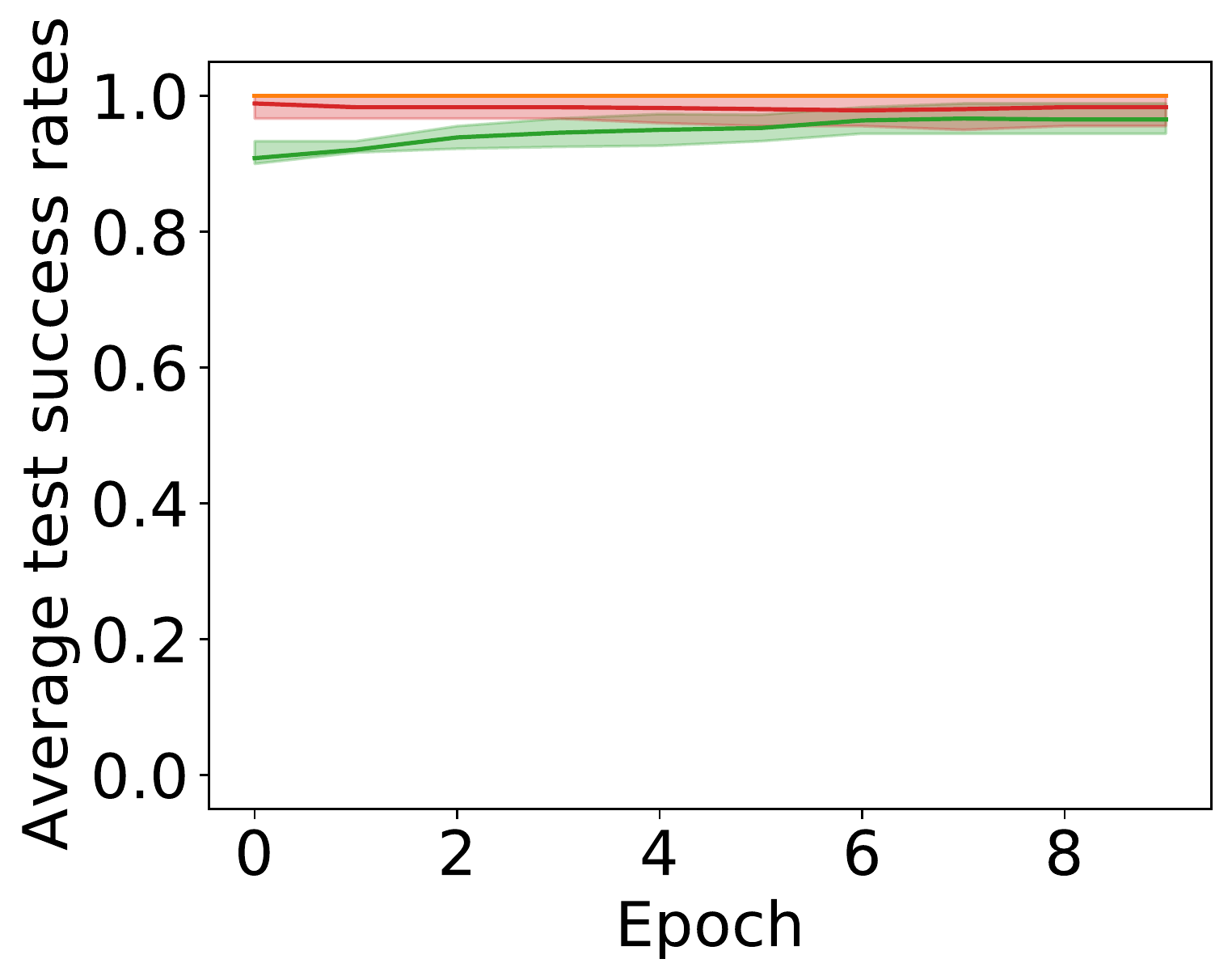}}
    \hfil
    \subfloat[\label{fig:pick-and-place}PickAndPlace]{
    \includegraphics[height=0.2\textwidth]{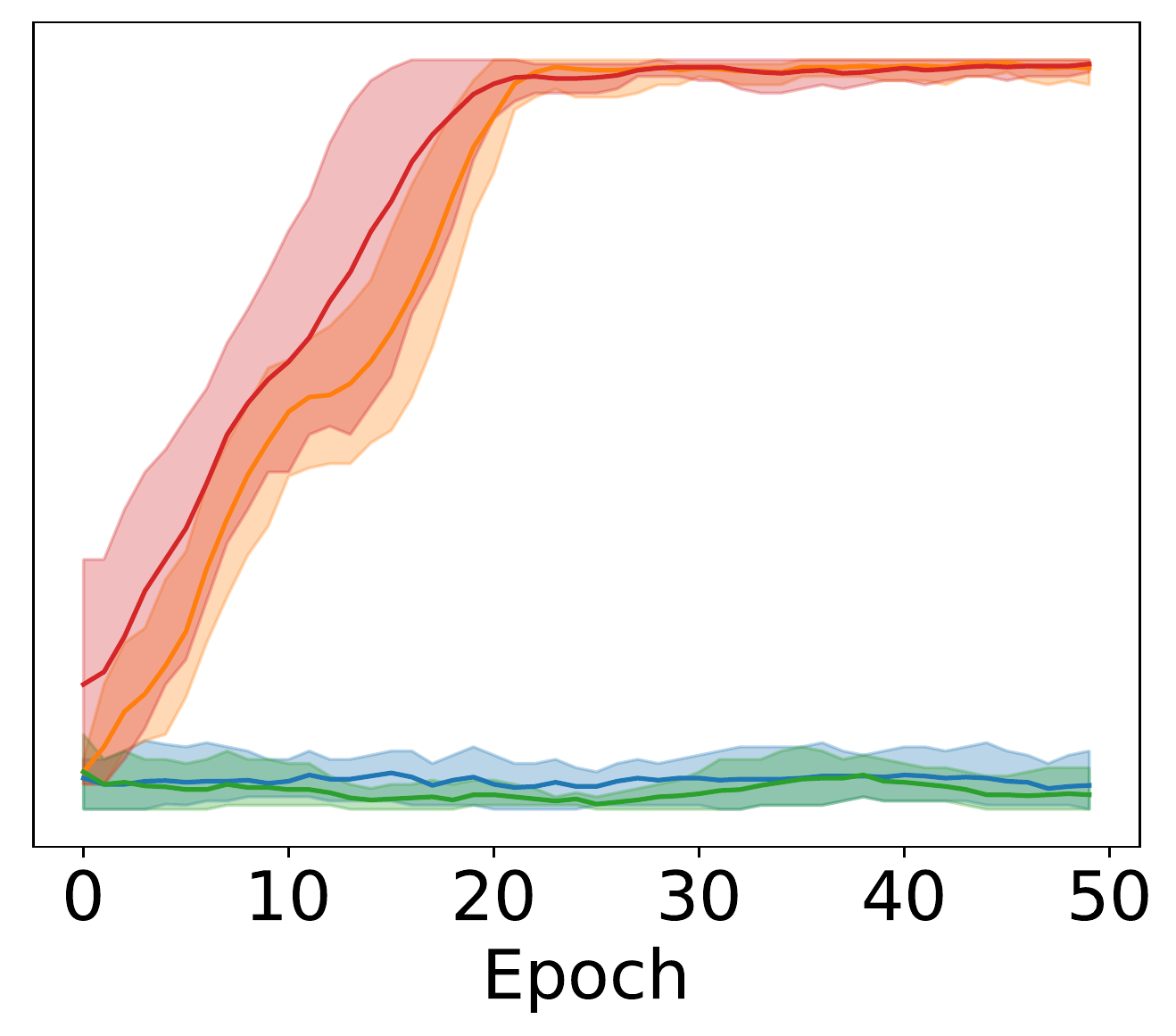}}
	\hfil
	\subfloat[\label{fig:slide}Slide]{
	\includegraphics[height=0.2\textwidth]{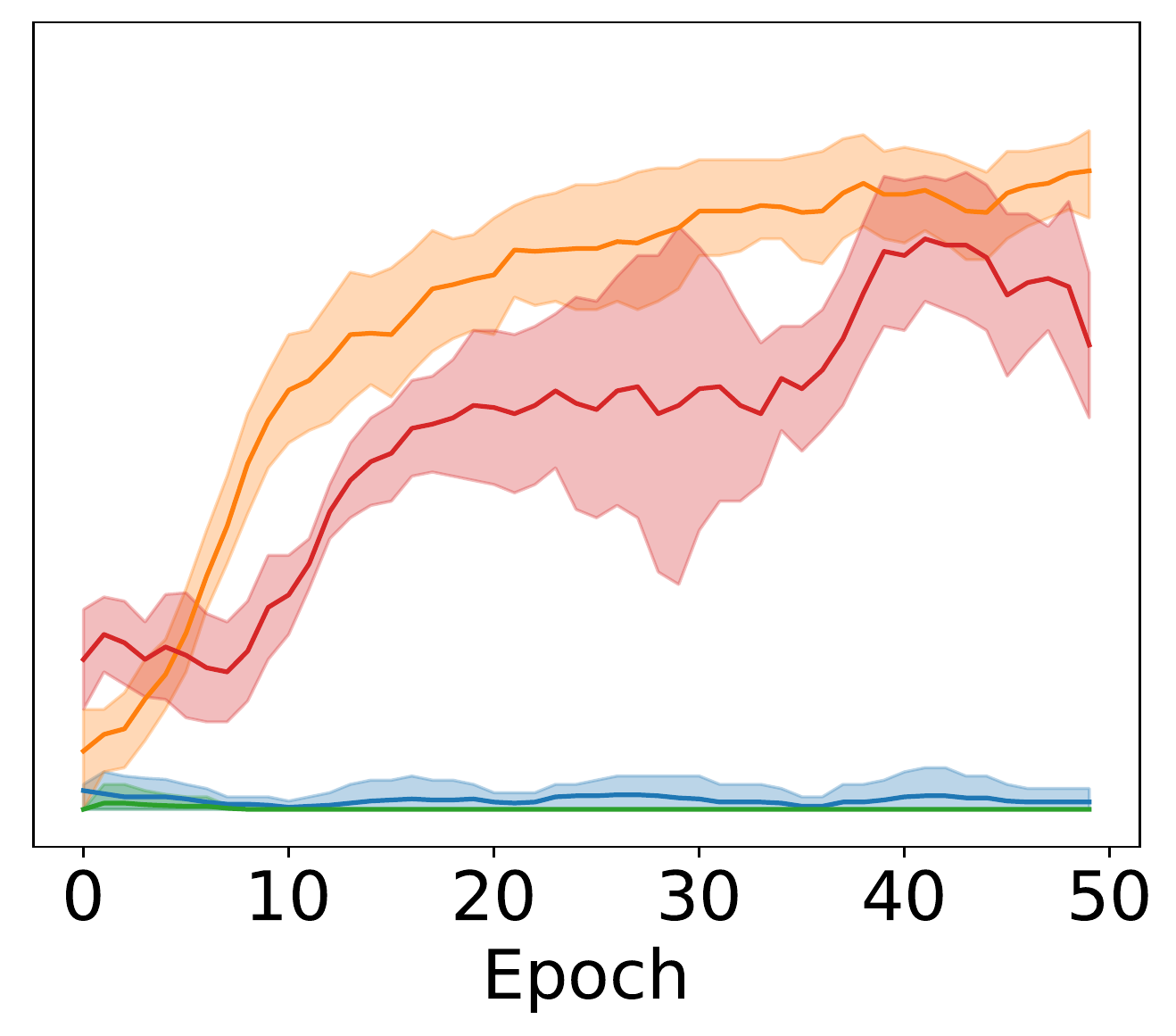}}
    \hfil
    \subfloat[\label{fig:push}Push]{
    \includegraphics[height=0.2\textwidth]{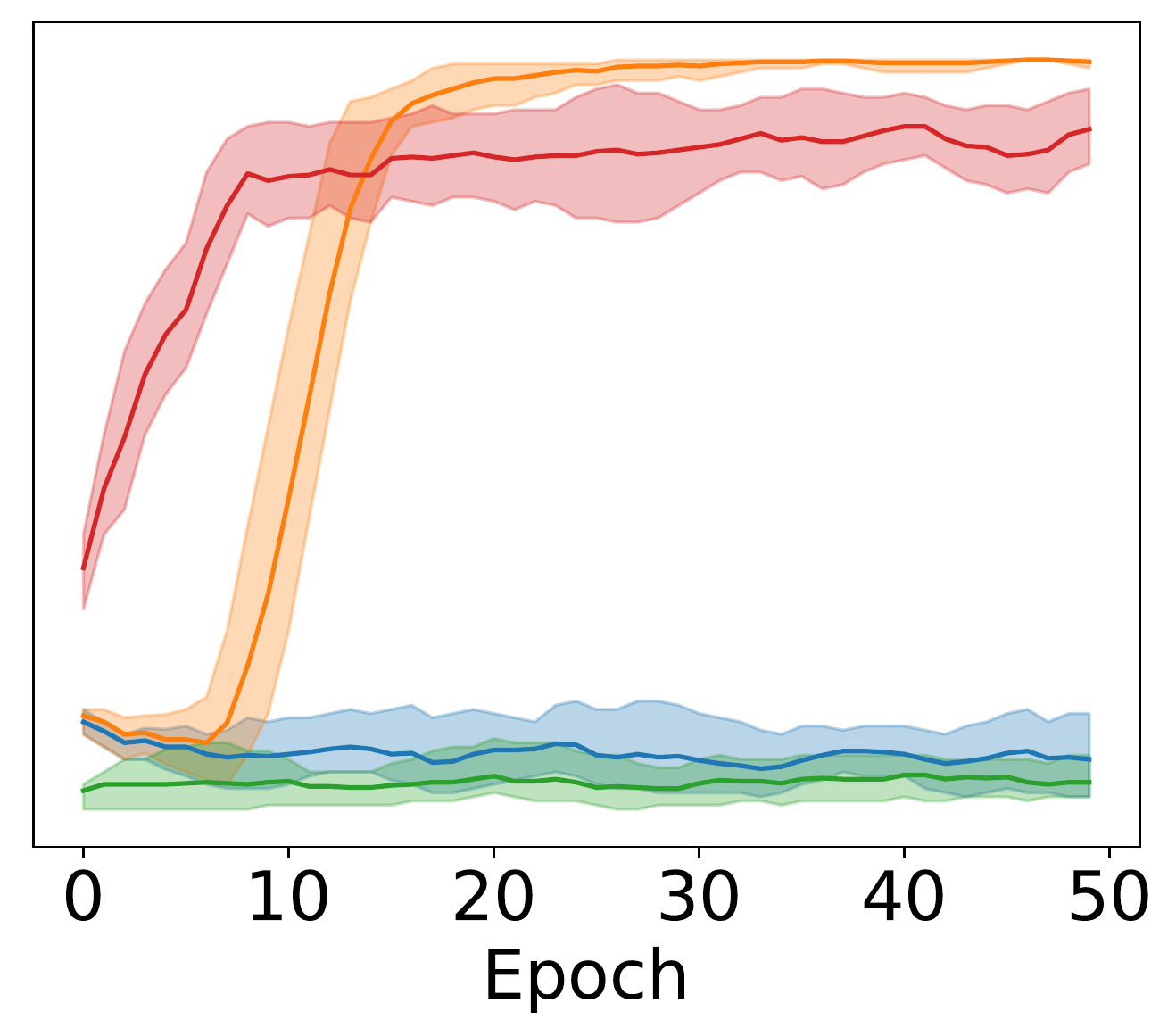}}
	\caption{Test success rates of the \textsf{single-step tasks} on Mujoco or Pybullet engine.\label{fig:single-step-task-result}}
	\vspace{-15pt}
\end{figure}

For comparison, we ran the same DDPG-HER algorithm \cite{Andrychowicz2017} with the same set of hyperparameters on the Mujoco- and our Pybullet-based environments. As shown in Fig.~\ref{fig:single-step-task-result}, the algorithm achieved almost the same performances on the both software environments
\footnote{Note that the \textsf{Slide} task is sensitive to the random seeds in both environments. The agent was unable to learn anything in some cases. It also exhibited higher variance than other tasks.}.
These results demonstrate our successful transplantation of the \textsf{single-step tasks} onto the Pybullet engine. 

Beside the original tasks, we also ran the experiments with joint space control using the same algorithm. These joint space control tasks differ from the original gripper frame control tasks in that the robot's actions are now joint commands, and the state representation further includes the current joint states. Results show that, in comparison to gripper frame control mode, \textsf{single-step tasks} under joint space control mode are harder to learn (Fig.~\ref{fig:joint-ctrl}). Its performance on the easiest \textsf{Reach} task also shows higher variance. 

This is expected as the action space has higher dimensionality. On the other hand, the gripper is constrained to be pointing top-down under the gripper frame control mode, but this constraint is released under the joint space control mode. This makes the tasks harder to learn by increasing the size of its solution space. 

For future research, it is valuable to develop reinforcement learning algorithms that can handle such control tasks with higher action dimensionality and larger solution space, potentially from (depth-) image observations. Investigating harder tasks including collision avoidance and comparing with classic motion planning methods are interesting directions as well.

\subsection{Benchmarking Multi-step tasks}\label{sec:multi-step-result}

This section discusses the performances of the DDPG-HER agent \cite{Andrychowicz2017} on the \textsf{multi-step tasks}, with and without the use of the proposed simplistic curriculum (\href{https://github.com/IanYangChina/taros2021codes/blob/master/taros20210308_supplementary.pdf}{supplementary material} section 3). We benchmarked the tasks without a chest using $2, 3, 4$ blocks, and the tasks with a chest using $1, 2, 3$ blocks.

We made one modification to the agent for these tasks. The action values predicted by the critic network are clipped within $[-50, 0]$ in the \textsf{single-step tasks} as suggested by \cite{Andrychowicz2017}, because the lowest value is $-50$ under sparse reward setting, given that the maximum episode timestep is $50$ \cite{plappert2018multi}. For the \textsf{multi-step tasks}, we changed the lower bound of the clipped value range to the negative maximum episode timestep for each task.

\begin{figure}[h]
	\vspace{-15pt}
    \centering
	\subfloat[\label{fig:blcok-rearrange}BlockRearrange]{
	\includegraphics[height=0.2\textwidth]{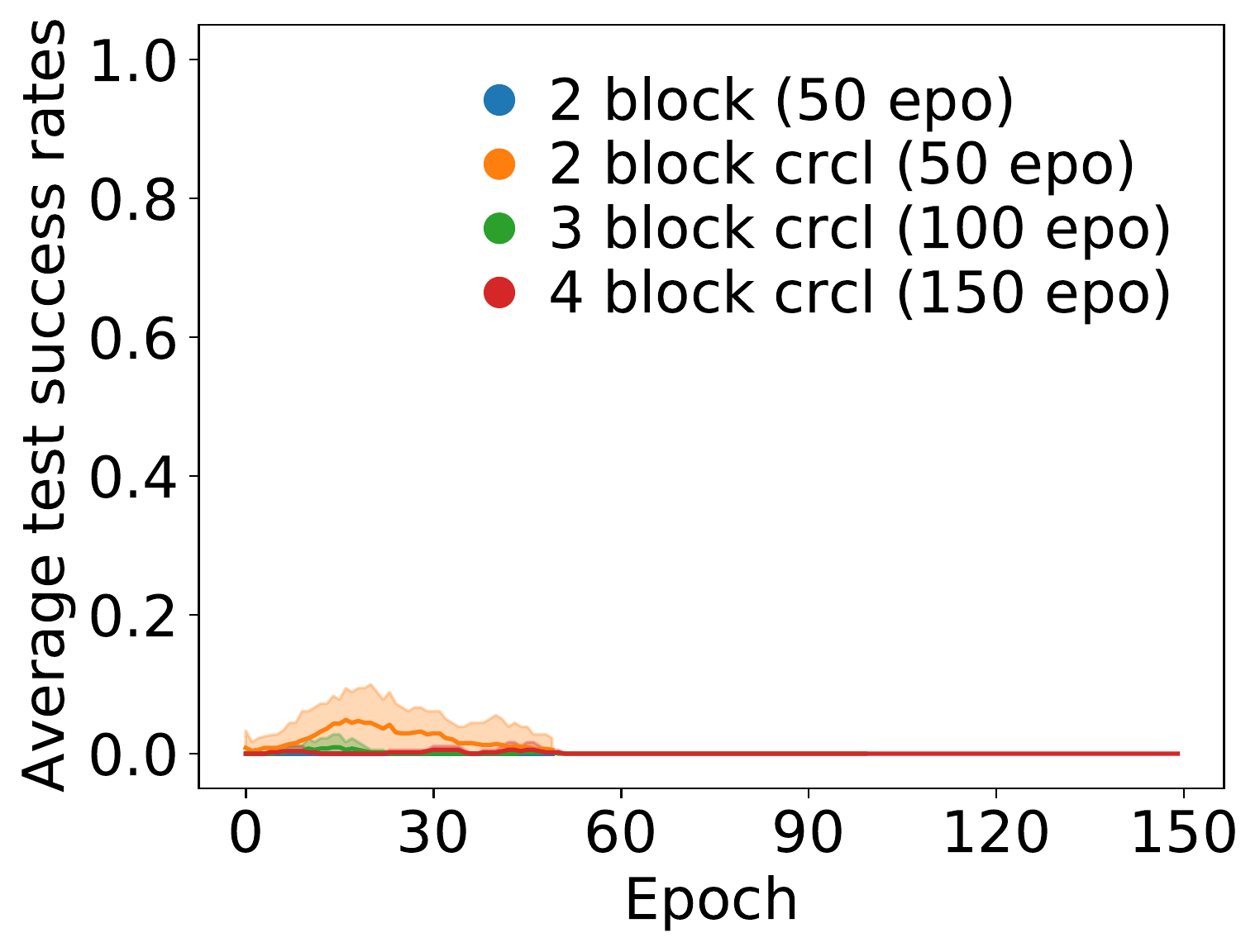}}
    \hfil
    \subfloat[\label{fig:chest-push}ChestPush]{
    \includegraphics[height=0.2\textwidth]{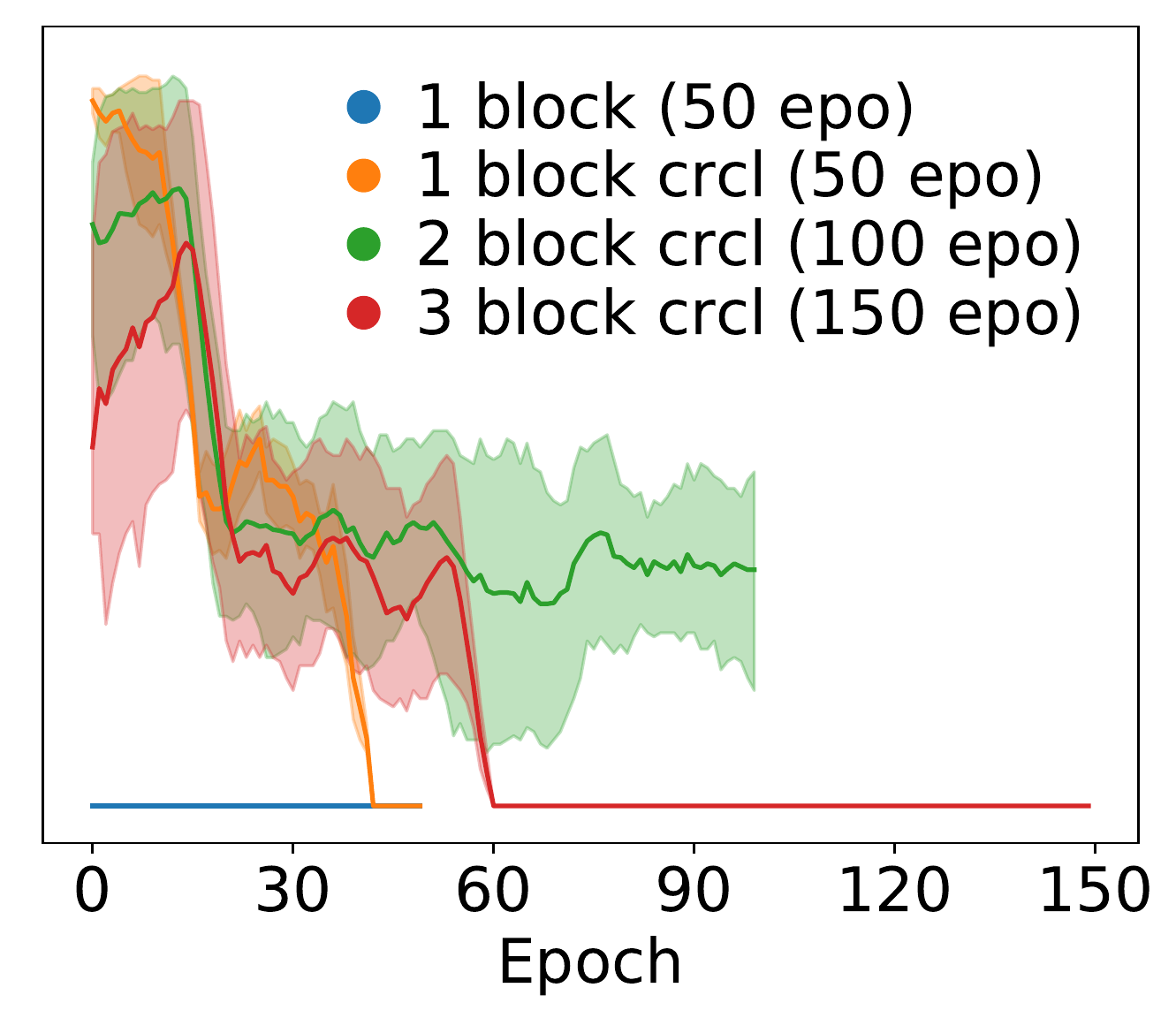}}
	\hfil
	\subfloat[\label{fig:chest-pick}ChestPick]{
	\includegraphics[height=0.2\textwidth]{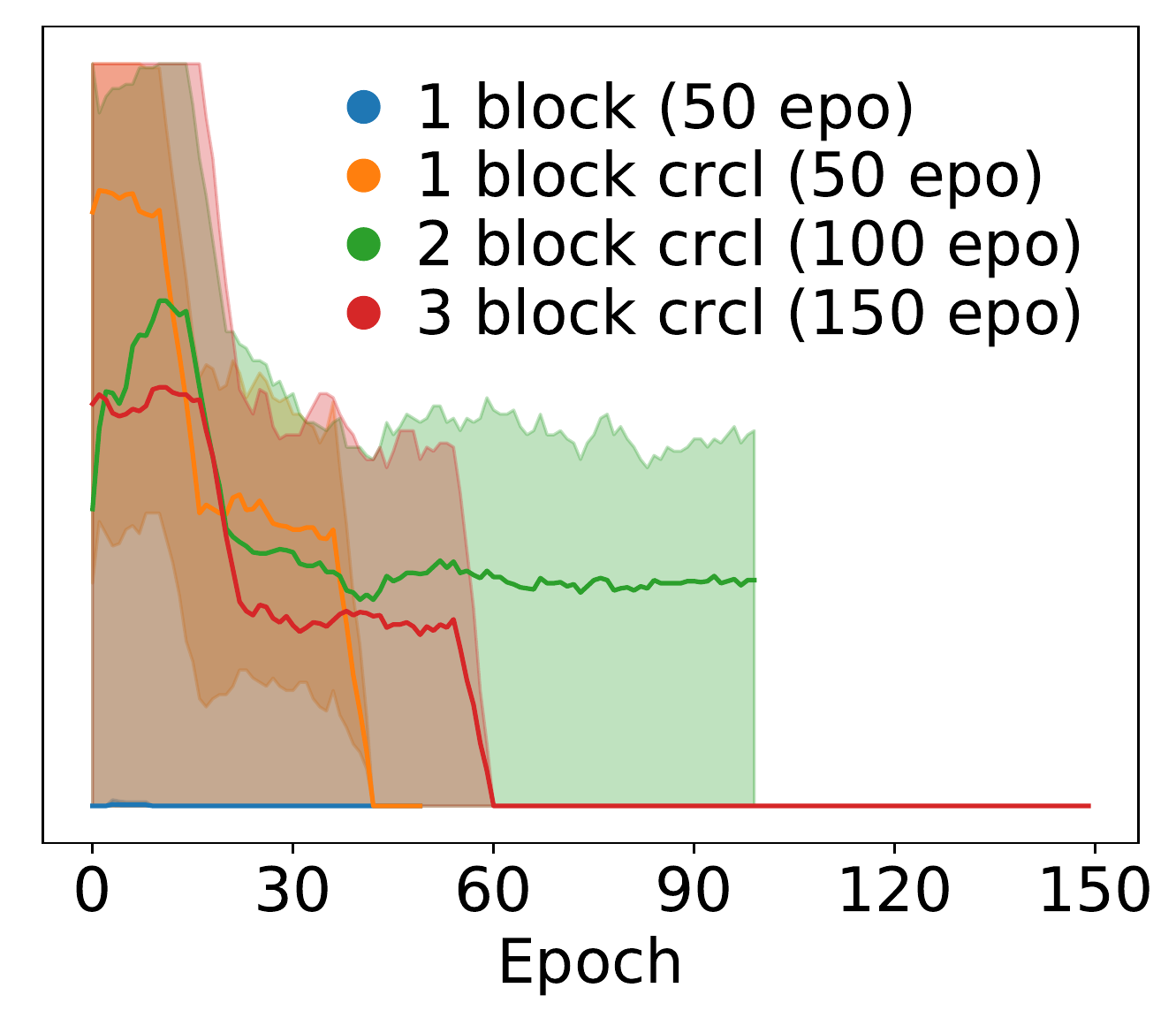}}
    \hfil
    \subfloat[\label{fig:block-stack}BlockStack]{
    \includegraphics[height=0.2\textwidth]{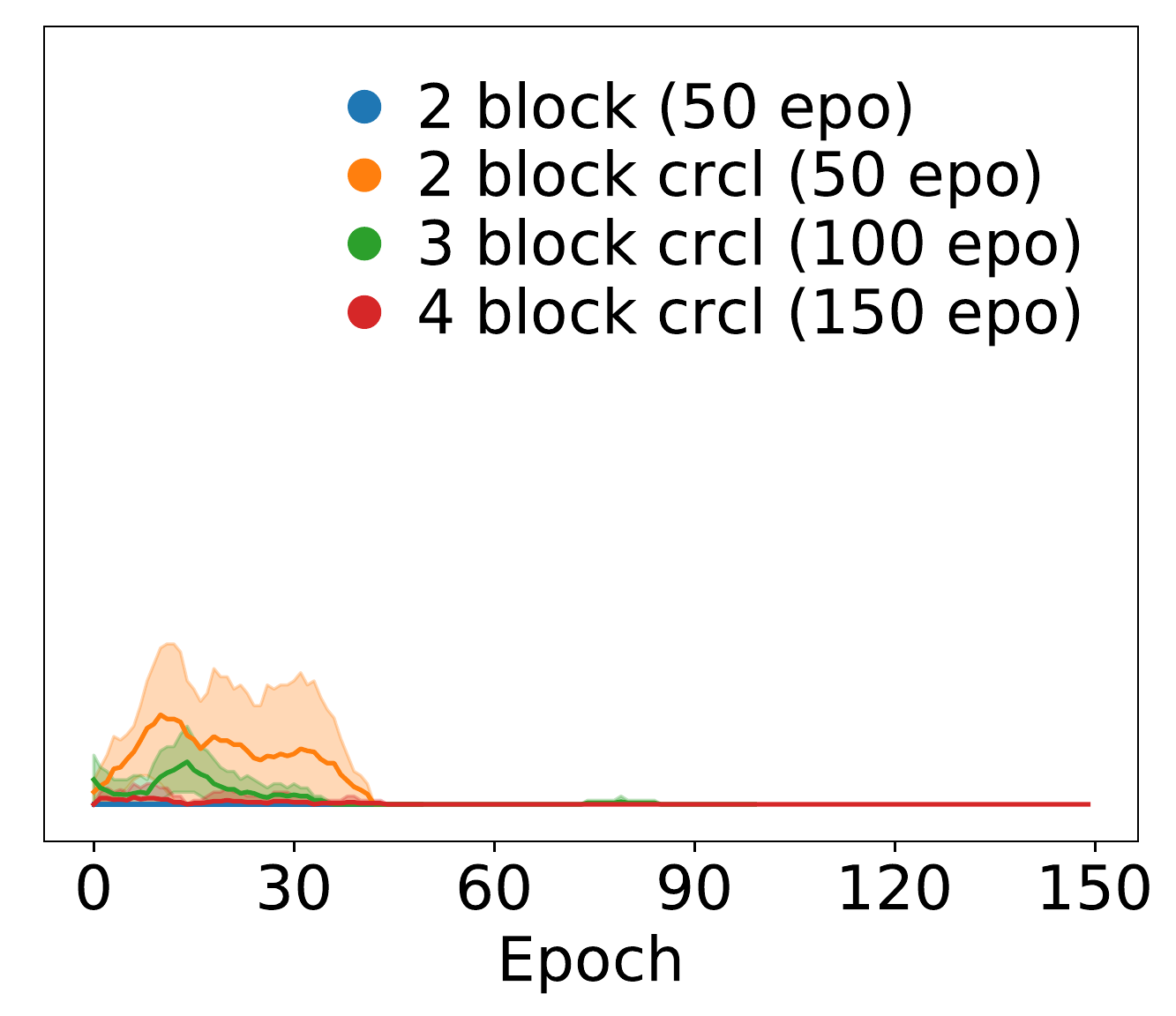}}
	\caption{Test success rates of the \textsf{multi-step tasks}. `crcl' means `curriculum'.\label{fig:multi-step-task-result}}
	\vspace{-15pt}
\end{figure}

As shown in Fig.~\ref{fig:multi-step-task-result}, the DDPG-HER agent learned nothing without the help of the curriculum (blue line in each subplot). When aided by the curriculum, it could achieve the easiest steps (open the chest door) in the \textsf{ChestPush} and \textsf{ChestPickAndPlace} tasks, but failed at later harder steps (success rates quickly drop to near $0$ as learning proceeds, shown by the orange, green and red lines). For the \textsf{BlockRearrange} and \textsf{BlockStack} tasks, the agent struggled to learn the easiest steps even with the help of the curriculum. This is because exploring to open the chest door is easier than moving a block around. These results indicate that these sparse rewards \textsf{multi-step tasks} are still unsolvable given the current state-of-the-art reinforcement learning algorithms.

\subsection{Challenges and opportunities}\label{sec:challenges}

This section discusses the challenges and future research opportunities related to the sparse reward multi-step robotic manipulation tasks from two perspectives, including exploration efficiency and representation learning. From each of them, there are several research directions that can be focused on.

\textbf{Exploration}: In sparse reward environments, improving exploration efficiency has long been a research challenge in the field of DRL \cite{rashid2020optimistic}. However, current research has been restricted within toy problems (e.g., grid world) or the Atari games (e.g., Montezuma's Revenge). These are all 2D tasks with discrete action spaces. Robotic manipulations are tasks in a 3D world, with larger and richer observations and continuous action spaces. It would be valuable to evaluate techniques that work in the 2D tasks on our 3D and continuous action tasks, with the hope to improve them further and transfer to the real-world. 

In the multi-goal setting, we have demonstrated the insufficiency of the HER aided by a simplistic goal generation curriculum. It is then potentially fruitful to develop a better curriculum for such tasks. Another interesting direction is to leverage task decomposition for multi-step tasks and make use of hierarchical learning systems \cite{yang2021hierarchical}. The use of sub-goals is a promising way to tackle the hard exploration problem in such tasks.

\textbf{Representation learning}: Representation for RL agents, especially in sparse reward tasks, has been increasingly active recently. Different from supervised learning tasks, RL agents rely on the reward signals to learn a representation of the environment and the task altogether. This makes it hard to generate and maintain a good representation in sparse reward tasks, in which the representation learnt can easily collapse. Again, current state-of-the-art in this direction has been largely restricted within 2D tasks or tasks with short horizon \cite{laskin2020curl,lyle2021effect}, and our environment is a promising testbed for evaluating and improving them in a 3D world with longer task horizons.

\section{Conclusion}\label{sec:conclusion}

We propose an open-source robotic manipulation simulation software implementation for multi-goal multi-step deep reinforcement learning. The implementation
of the OpenAI multi-goal\-styled environment (based on the Mujoco engine) has been achieved using Pybullet. Performance of the popular DDPG-HER algorithm has been reproduced in our work (section~\ref{sec:single-step-result}). Except for the original manipulation tasks, named \textsf{single-step tasks}, we designed a set of \textsf{multi-step tasks} with sparse rewards in longer task horizons. We benchmarked the performances of the DDPG-HER agent with and without the use of a simplistic goal generation curriculum (section~\ref{sec:multi-step-result}), demonstrating the inability of the state-of-the-art algorithms to learn in such long horizon and sparse reward environments. Finally, we provided brief discussions of the challenges and future research opportunities, including \textsc{exploration} and \textsc{representation learning} in sparse reward reinforcement learning. Our future research will focus on developing sub-goal-based solutions to tackle such multi-step sparse reward robotic manipulation tasks.

\bibliographystyle{splncs04}
\bibliography{taros_arxiv}

\title{An Open-Source Multi-Goal Reinforcement Learning Environment for Robotic Manipulation with Pybullet (Supplementary)\thanks{The authors thank the China Scholarship Council (CSC) for financially supporting Xintong Yang in his PhD programme (No. 201908440400).}}

\titlerunning{This paper was submitted to Taros 2021}

\author{Xintong Yang\inst{1}\orcidID{0000-0002-7612-614X} \and
Ze Ji\inst{1}\orcidID{0000-0002-8968-9902}\thanks{Corresponding author} \and
Jing Wu\inst{2}\orcidID{0000-0001-5123-9861} \and
Yu-Kun Lai\inst{2}\orcidID{0000-0002-2094-5680}}

\authorrunning{X. Yang et al.}

\institute{Centre for Artificial Intelligence, Robotics and Human-Machine Systems (IROHMS), School of Engineering, Cardiff University, Cardiff, UK \\\email{\{yangx66, jiz1\}@cardiff.ac.uk} \and
School of Computer Science and Informatics, Cardiff University, Cardiff, UK \\\email{\{wuj11, laiy4\}@cardiff.ac.uk}}

\maketitle

\section{Multi-step tasks details}\label{apex:multi-step-task-detail}

\begin{figure}[h]
    \centering	
	\subfloat[\label{fig:init-state}Initial state distribution]{
	\includegraphics[height=0.39\textwidth]{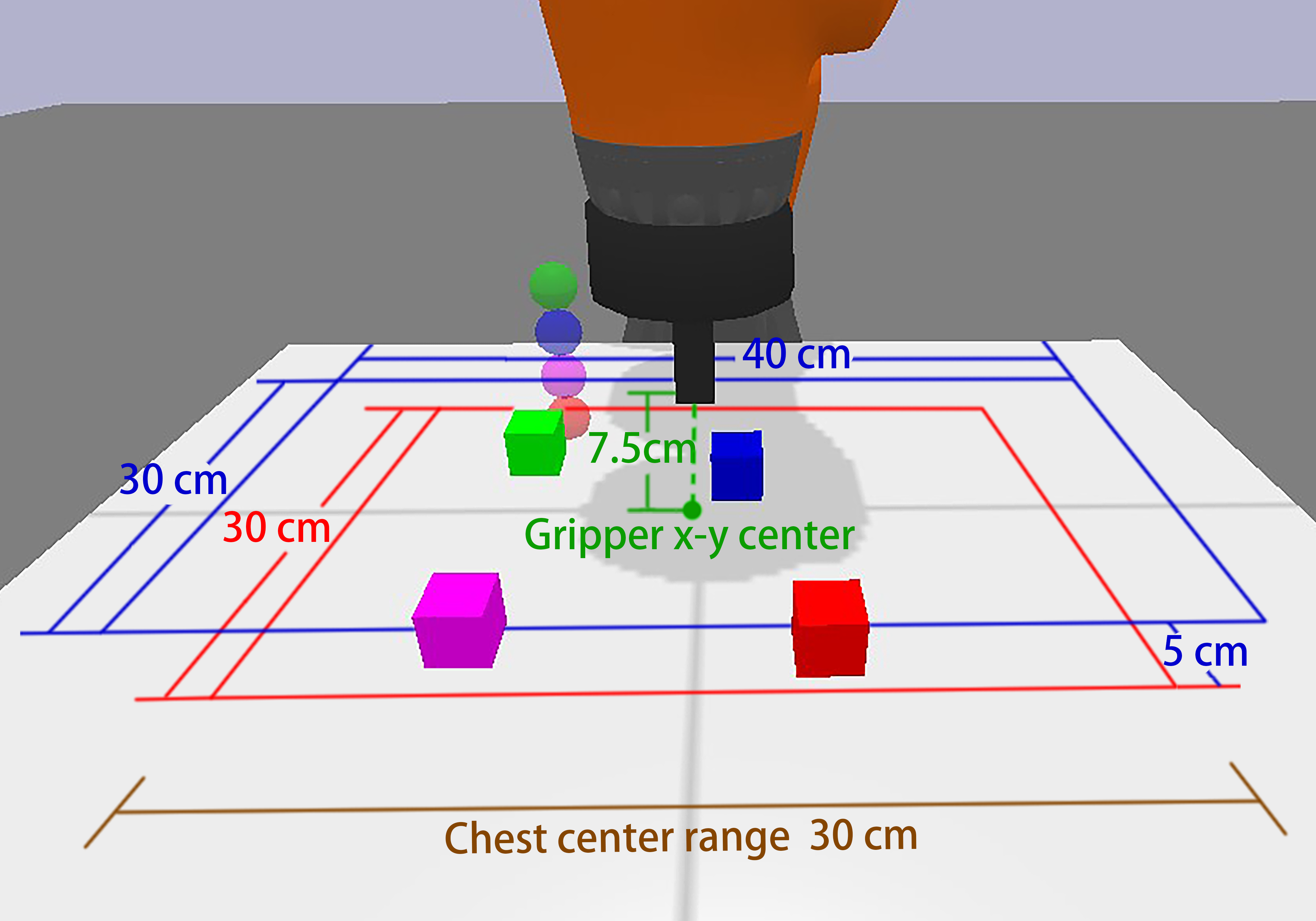}}
    \hfil
    \subfloat[\label{fig:chest-state}Chest state representation]{
    \includegraphics[height=0.39\textwidth]{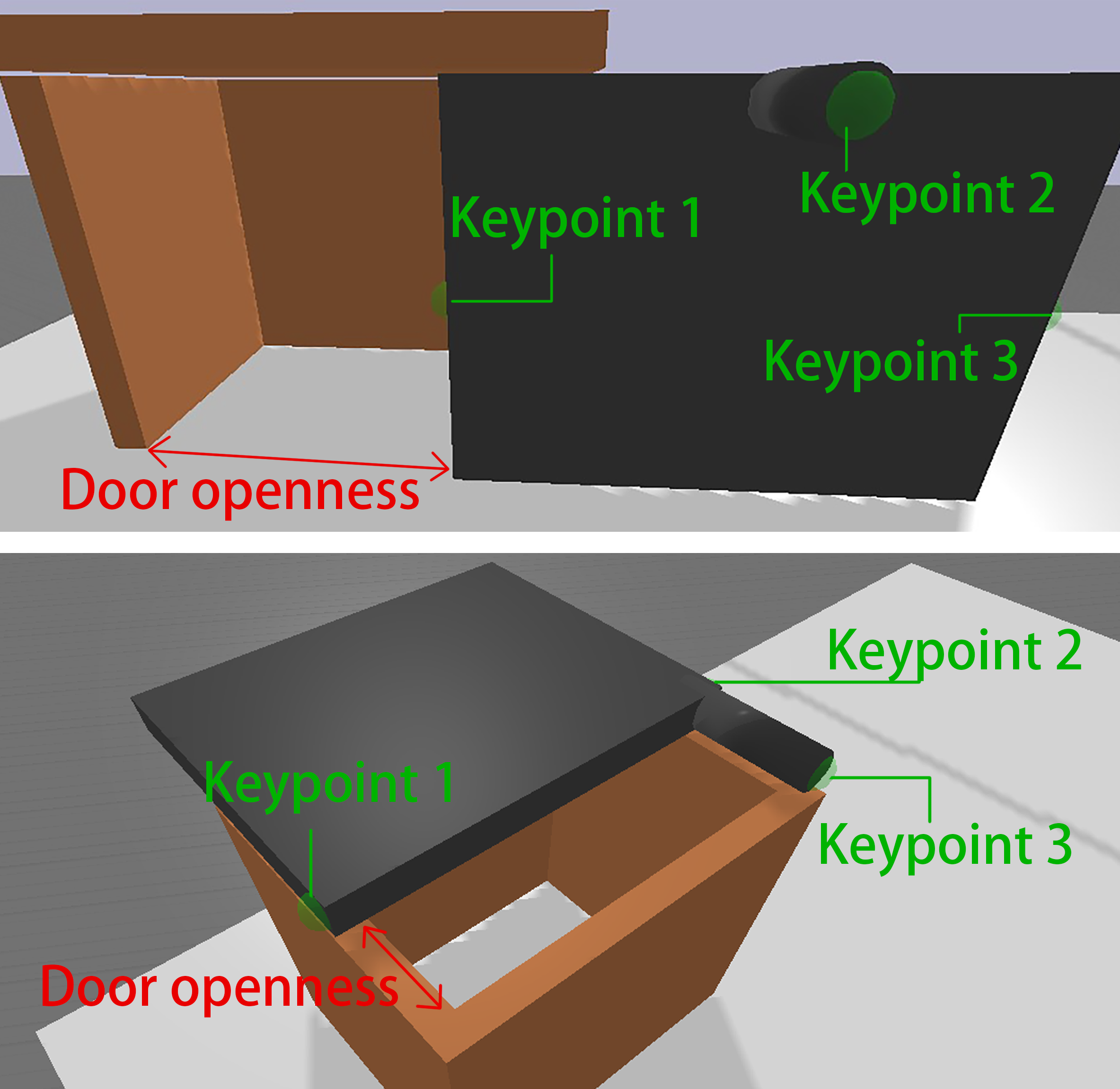}}
	\caption{(a) Initial state distribution. \textcolor{red}{Red line}: initial block position range for tasks without a chest. \textcolor{blue}{Blue line}: initial block position range for tasks with a chest. \textcolor{Brown}{Brown line}: initial position range for the centre of the chest. \textcolor{green}{Green dot}: initial gripper tip position for the \textsf{BlockRearrange} and \textsf{ChestPush} tasks, the point $0.075$ above is the initial position for the other two tasks. (b) Chest state representation: the Cartesian coordinates of the three keypoints (\textcolor{green}{green dots}) and the width of the door gap (\textcolor{red}{red line}). One keypoint is at the end of the door handle, the other two are on the two edges of the door. \label{fig:task-representation}}
\end{figure}

\textbf{Initial state distribution}\footnote{All length values used in the environment are in metres, angles in radians, unless otherwise specified.}: 
Fig~\ref{fig:init-state} gives a visualisation. For the \textsf{block\_stack} and \textsf{block\_rearrange} tasks, blocks' positions are sampled randomly within a square of $0.03$ width, centred at the gripper's $x$-$y$ coordinates (red line). For the \textsf{chest\_push} and \textsf{chest\_pick\_and\_place} tasks, they are sampled within a rectangle of length $0.04$ and width $0.03$, centred at the location $0.05$ away from the gripper's $x$-$y$ coordinates, nearer to the robot base (blue line). The position of the chest is randomly sampled on a $0.03$ line, $0.07$ away from the robot base (brown line). The initial pose of the gripper tip frame is fixed. For pushing tasks, it is at the centre of the table surface, while for picking tasks, it is $0.075$ above the table centre (green line). The difference of the initial gripper tip location is for easier exploration, a design inherited from the OpenAI Gym multi-goal tasks\footnote{This was not specified in the paper, but we found it in the \href{https://github.com/openai/gym/tree/master/gym/envs/robotics/fetch}{source codes of the package}.}.

\textbf{State representation}: All tasks share the same state representation for the Kuka robot and $n$ blocks, $\mathbf{s} = \{\mathbf{s}_{robot}||\mathbf{s}_{block\_1}||...||\mathbf{s}_{block\_n}\}$ (`$||$' denotes vector concatenation). The robot state includes the Cartesian coordinates and linear velocity of the gripper tip in the world frame, the finger width and velocity in the world frame. That is 

\[
\mathbf{s}_{robot} = (x_{grip}, y_{grip}, z_{grip}, vx_{grip}, vy_{grip}, vz_{grip}, w_{finger}, v_{finger})
\]

The state of each joint is only included if joint space control is used. A block state includes its Cartesian coordinates and Euler orientation angles in the world frame, its relative position, relative linear velocities and angular velocities with respect to the gripper tip. That is (for the $n$-th block),

\begin{equation*}
\begin{split}
\mathbf{s}_{block\_n} = ( & x_{block\_n}, y_{block\_n}, z_{block\_n}, roll_{block\_n}, pitch_{block\_n}, yaw_{block\_n}, \\
& x_{block\_n}^{rel}, y_{block\_n}^{rel}, z_{block\_n}^{rel}, vx_{block\_n}^{rel}, vy_{block\_n}^{rel}, vz_{block\_n}^{rel}, \\
& vroll_{block\_n}^{rel}, vpitch_{block\_n}^{rel}, vyaw_{block\_n}^{rel})
\end{split}
\end{equation*}

For tasks involving a chest, the chest state includes the Cartesian coordinates of the three green keypoints attached on the door and how wide the door is opened (see Fig.~\ref{fig:chest-state}). 

\textbf{Action space}: All actions are multi-dimensional and continuous in $[-1, 1]$ as a common definition for continuous control reinforcement learning. For Cartesian space control, an action is mapped to the changes of the gripper tip coordinates in the world frame by multiplying with $0.05$\footnote{An action of $1.0$ at the $x$ element will move the gripper towards the positive $x$ direction for $0.05$.}. 
For joint space control, an action is mapped to the changes of the joint angle by multiplying with $0.05$. The finger width control dimension is mapped to the range of the finger state, such that $-1$ corresponds to the fingers being fully opened (symmetric fingers). Only the \textsf{ChestPickAndPlace} and \textsf{BlockStack} tasks require gripper finger control. In Cartesian space control mode, the gripper tip movement is bounded within a box of length $0.4$, width $0.3$ and height $0.375$, placed on the table surface centre. In joint space control mode, the robot arm movement is bounded by the joint limits.

\textbf{Goal representation and generation:} A goal consists of the Cartesian coordinates of all the $n$ blocks in the world frame. If a chest is involved, the goal includes an extra scalar indicating the largest openness of the door. That is

\[
\mathbf{g} = (
x_{block\_1}, y_{block\_1}, z_{block\_1}, ..., x_{block\_n}, y_{block\_n}, z_{block\_n})
\]

A target is constrained to not overlap with the blocks and other targets given a threshold ($0.06$ by default).

\begin{itemize}
\item[$\bullet$] \textsf{BlockRearrange}: Target block positions (desired goals) are sampled within the same square (red line in Fig.~\ref{fig:init-state}) in which the initial block positions are sampled.

\item[$\bullet$] \textsf{ChestPush}: Target block positions are fixed at the centre of the chest, on the table surface. With a goal achieving distance threshold of $0.1$, this means the task is to push the blocks into a sphere of radius $0.05$ centred at the chest centre on the table.

\item[$\bullet$] \textsf{ChestPickAndPlace}: This task is the same as the \textsf{ChestPush} task, except that the robot needs to pick and drop the blocks, rather than push.

\item[$\bullet$] \textsf{BlockStack}: This task first samples a random order in which the blocks need to be stacked, then samples a tower position within the initial block position square (red line in Fig.~\ref{fig:init-state}).
\end{itemize} 

\textbf{Reward function}: Every task comes with a dense and a sparse reward function, based on a desired and an achieved goal. The dense reward function outputs the negative Euclidean distance of the two goals. That is 

\[r_{dense} = -||\mathbf{g}_{achieved}-\mathbf{g}_{desired}||_2\]

The sparse reward function outputs $0$ when a desired goal is achieved and $-1$ otherwise. A goal is regarded as achieved when the Euclidean distance of the two goals is smaller than or equal to a given threshold $\delta$, ($\delta=0.05$ by default). That is

\begin{equation}
  r_{sparse} =
    \begin{cases}
      0 & ||\mathbf{g}_{achieved}-\mathbf{g}_{desired}||_2 <= \delta\\
      -1 & \text{otherwise}
    \end{cases}       
\end{equation}

\textbf{Task horizon}: The number of timesteps differs based on the number of blocks involved in a task. Every task has a base number of $50$ timesteps, and adding one block increases it by $25$. For example, a task of stacking two blocks and a task of pushing one block into a chest both provide a task horizon of $75$ timesteps.

\section{Curriculum details}\label{apex:curriculum}
The simple curriculum used in (main text) section 3.2 starts by generating easy goals and gradually increases difficulty. It first computes the total number of goals to be generated in the whole training process, which effectively equals the total number of episodes. It then separates the total number of training goals evenly into a number of  difficulty levels. The difficulty levels for each task are defined as follows.

\begin{itemize}
\item[$\bullet$] \textsf{BlockRearrange}: The number of levels is equal to the number of blocks. The easiest one is to generate only one random target position, and other target positions are made equal to the blocks' positions. In other words, the curriculum starts by asking the robot to push one block, and gradually increases the number of blocks to push.

\item[$\bullet$] \textsf{ChestPush}: The number of levels is equal to the number of blocks plus one. The easiest level is then to open the door of the chest. The following level starts by pushing one block, and increases to all the blocks.

\item[$\bullet$] \textsf{ChestPickAndPlace}: The same as the \textsf{ChestPush} task, except that the robot is asked to pick and drop the blocks, rather than push.

\item[$\bullet$] \textsf{BlockStack}: The number of levels is equal to the number of blocks. The easiest one is to push or pick-and-place the base block to a random position, and other blocks stay unmoved. Each of the following levels then adds one more block to be stacked.
\end{itemize}

For each episode, the method samples a level of difficulty for the goal to generate. It maintains a record of the number of generated goals from each level. At the beginning of a training process, the easiest level has a sampling probability of $1.0$, and other levels have $0$ chances. When the number of generated goals of a level passes half the required number of goals to generate, the method sets the probabilities of this level and its next level to $0.5$. When a level finishes generating all the goals, its probability is set to $0$ and its next level's probability is set to $1.0$ if this next level has not passed half the number of goals to generate. The whole process repeats until the last difficulty level is finished, which means the training process finishes. 

To reduce unnecessary training time, the task horizon changes based on the current curriculum level. At the lowest level the task has $50$ episode timesteps, and going one level up increases it by $25$ timesteps. This means the task horizon increases as the agent is given harder and harder goals to achieve.

This method is totally based on human prior and rather simplistic. As demonstrated by the results (main text section 3.2), it does help the algorithm to learn at the beginning from easier goals, but it consistently fails as goals become more difficult. This indicates that a naive curriculum alone is not enough to achieve such long horizon multi-step manipulation tasks in an extremely sparse reward setting.

\section{API details}\label{apex:APIs-detail}
This section explains the meanings of the arguments of the \texttt{make\_env(...)} function in more details. Table~\ref{tab:make-env-arg-meaning} illustrates these meanings. Table~\ref{tab:tasks} lists all the strings taken by the \texttt{task} argument and their corresponding tasks. \hyperref[code:camera-setup]{Code 2} provides an example of setting up camera parameters for rendering observations and goal images.

\begin{table}[h]
\caption{Meanings and data types of the \texttt{make\_env} function arguments\label{tab:make-env-arg-meaning}}
\begin{center}
\begin{tabularx}{\textwidth}{llX}
\toprule
Argument & Type & Meaning\\
\hline
\texttt{task} & String & The name of the task environment to create.\\
\texttt{joint\_control} & Boolean & Whether to use joint control actions.\\
\texttt{num\_block} & Integer & The number of blocks involved. Only used in \textsf{multi-step tasks}.\\
\texttt{render} & Boolean & Whether to create a GUI window rendering the simulation.\\
\texttt{binary\_reward} & Boolean & Whether to use sparse reward signals.\\
\texttt{max\_episode\_step} & Integer & The number of timesteps of an episode.\\
\texttt{distance\_threshold} & Float & The threshold used to determine whether a goal is achieved.\\
\texttt{image\_observation} & Boolean & Whether to use images as observations.\\
\texttt{Depth\_image} & Boolean & Whether to include depth information in images.\\
\texttt{goal\_image} & Boolean & Whether to use images as goals.\\
\texttt{visualize\_target} & Boolean & Whether to render a semitransparent sphere at the target position of the manipulated objects.\\
\texttt{camera\_setup} & List of Dict & A list of dictionary contains camera parameters, please see \hyperref[code:camera-setup]{Code 2} for an example.\\
\texttt{observation\_cam\_id} & Integer & The index of the dictionary in the \texttt{camera\_setup} list used to render observation images.\\
\texttt{goal\_cam\_id} & Integer & The index of the dictionary in the \texttt{camera\_setup} list used to render goal images.\\
\texttt{use\_curriculum} & Boolean & Whether to use the simple curriculum goal generation method, see section~\ref{apex:curriculum}. Only used in \textsf{multi-step tasks}.\\
\texttt{num\_goal\_to\_generate} & Integer & The number of goals to be generated in the whole training process, normally equal to the total number of training episodes. Only used in \textsf{multi-step tasks}.\\
\bottomrule
\end{tabularx}
\end{center}
\end{table}

\begin{table}[h]
\caption{Correspondences between the \texttt{task} argument and each task\label{tab:tasks}}
\begin{center}
\begin{tabular}{lccl}
\toprule
\texttt{task=} & & & task\\
\hline
`reach' & & & KukaReach\\
`push' & & & KukaPush\\
`slide' & & & KukaSlide\\
`pick\_and\_place' & & & KukaPickAndPlace\\
`block\_stack' & & & BlockStack\\
`block\_rearrange' & & & BlockRearrange\\
`chest\_pick\_and\_place' & & & ChestPickAndPlace\\
`chest\_push' & & & ChestPush\\
\bottomrule
\end{tabular}
\end{center}
\end{table}

\begin{table}[h]
\begin{center}
\begin{tabularx}{\textwidth}{lX}
\toprule
\textbf{Code 2}\label{code:camera-setup} A list of camera setup dictionary & Meaning\\
\midrule
camera\_setup = [ &\\
\ \ \ \ \ \ \{ &\\
\ \ \ \ \ \ \ \ \ \ \ \ 'cameraEyePosition': [-1.0, 0.25, 0.6], & the 3D coordinates of the camera frame in the world frame\\
\ \ \ \ \ \ \ \ \ \ \ \ 'cameraTargetPosition': [-0.6, 0.05, 0.2], & the 3D coordinates which the camera looks at in the world frame\\
\ \ \ \ \ \ \ \ \ \ \ \ 'render\_width': 128, & the width of the rendered image\\
\ \ \ \ \ \ \ \ \ \ \ \ 'render\_height': 128 & the height of the rendered image\\
\ \ \ \ \ \ \}, &\\
\ \ \ \ \ \ \{ &\\
\ \ \ \ \ \ \ \ \ \ \ \ 'cameraEyePosition': [-1.0, -0.25, 0.6], &\\
\ \ \ \ \ \ \ \ \ \ \ \ 'cameraTargetPosition': [-0.6, -0.05, 0.2], &\\
\ \ \ \ \ \ \ \ \ \ \ \ 'render\_width': 128, &\\
\ \ \ \ \ \ \ \ \ \ \ \ 'render\_height': 128 &\\
\ \ \ \ \ \ \} &\\
] &\\
\bottomrule
\end{tabularx}
\end{center}
\end{table}

\end{document}